\documentclass[sigconf]{acmart} 
\acmSubmissionID{5754}
\AtBeginDocument{%
  }

\setcopyright{acmlicensed}
\copyrightyear{2018}
\acmYear{2018}
\acmDOI{XXXXXXX.XXXXXXX}

\acmConference[Conference acronym 'XX]{Make sure to enter the correct
  conference title from your rights confirmation emai}{June 03--05,
  2018}{Woodstock, NY}
\acmISBN{978-1-4503-XXXX-X/18/06}

\usepackage{algorithm}
\usepackage{algorithmic}
\usepackage{subfigure}
\usepackage{hyperref}
\usepackage{multirow}
\usepackage{color}
\usepackage{colortbl}
\usepackage{caption}
\newcommand{\bs}[1]{\boldsymbol{#1}}

\begin{document}

\title{Simplifying Graph Kernels for Efficient}

\author{Lin Wang}
\affiliation{
    \institution{The Hong Kong Polytechnic University}
    \city{Hong Kong}
    \country{China}}
\email{comp-lin.wang@connect.polyu.hk}

\author{Shijie Wang}
\affiliation{
    \institution{The Hong Kong Polytechnic University}
    \city{Hong Kong}
    \country{China}}
\email{shijie.wang@connect.polyu.hk}

\author{Sirui Huang}
\affiliation{
    \institution{The Hong Kong Polytechnic University}
    \city{Hong Kong}
    \country{China}}
\email{sirui.huang@connect.polyu.hk}

\author{Qing Li}
\authornote{Corresponding author. \\ This paper is currently under review.}
\affiliation{
	\institution{The Hong Kong Polytechnic University}
 \city{Hong Kong}
 \country{China}
}
\email{csqli@comp.polyu.edu.hk}

\renewcommand{\shortauthors}{Trovato et al.}





\begin{abstract}

While kernel methods and Graph Neural Networks offer complementary strengths, integrating the two has posed challenges in efficiency and scalability. The Graph Neural Tangent Kernel provides a theoretical bridge by interpreting GNNs through the lens of neural tangent kernels. However, its reliance on deep, stacked layers introduces repeated computations that hinder performance.
In this work, we introduce a new perspective by designing the simplified graph kernel, which replaces deep layer stacking with a streamlined $K$-step message aggregation process. This formulation avoids iterative layer-wise propagation altogether, leading to a more concise and computationally efficient framework without sacrificing the expressive power needed for graph tasks.
Beyond this simplification, we propose another Simplified Graph Kernel, which draws from Gaussian Process theory to model infinite-width GNNs. Rather than simulating network depth, this kernel analytically computes kernel values based on the statistical behavior of nonlinear activations in the infinite limit. This eliminates the need for explicit architecture simulation, further reducing complexity.
Our experiments on standard graph and node classification benchmarks show that our methods achieve competitive accuracy while reducing runtime. This makes them practical alternatives for learning on graphs at scale. Full implementation and reproducibility materials are provided at:
\href{https://anonymous.4open.science/r/SGNK-1CE4/}{https://anonymous.4open.science/r/SGNK-1CE4/}.
\end{abstract}

\maketitle

\section{Introduction}


Graphs are collections of nodes and edges that represents relationships between entities in a structured format, widely used to model complex networks, such as social networks~\cite{chen2021social, tsao2021social}, recommender systems~\cite{he2020lightgcn, wang2024multi, wu2020diffnet++}, molecular analysis~\cite{wang2022molecular, rong2020self}, and transportation networks~\cite{velivckovic2023everything, diao2021impacts}.
Recently, the widespread application of graph-structured data has attracted significant attention to Graph Neural Networks (GNNs) and Graph Kernels (GKs)~\cite{wang2024graph, wei2023neural, kang2022lr}.

Typically, GNNs~\cite{waikhom2023survey, xu2018powerful} learn node or graph representations through recursive message passing and feature extraction. In this process, information is propagated from a node's neighbors, while linear transformations, combined with non-linear activations, extract node features. By stacking these operations, GNNs effectively capture higher-order information from neighboring nodes.
In contrast, GKs embed graph data into a high-dimensional feature space to measure similarities based on structural information and node features~\cite{vishwanathan2010graph, nikolentzos2021graph}. Unlike GNNs, GKs do not require training and are advantageous for small-scale graphs, making them suitable for scenarios with limited computational resources~\cite{vishwanathan2010graph}.
Despite their effectiveness, each method has distinct limitations. GNNs, as parameterized methods, require extensive training, making them sensitive to initialization and reliant on resource-intensive parameter tuning, which affects efficiency and scalability~\cite{zhou2020graph,wu2020comprehensive}. In contrast, GKs are non-parametric methods that leverage node features and structural information for similarity measurement. While GKs avoid intensive training, their representation capabilities are less robust than those of GNNs, which may lead to sub-optimal performance~\cite{shervashidze2011weisfeiler, xia2019random, borgwardt2005shortest}.



To harness the advantages of both GNNs and GKs, ~\citet{du2019graph} introduced the Graph Neural Tangent Kernel (GNTK), a novel kernel method that effectively bridges the gap between these two approaches. Specifically, the GNTK models the training dynamics of infinite-width GNNs, thereby integrating the powerful representational capabilities of GNNs with the computational efficiency of GKs. By capturing the strengths of both paradigms, the GNTK has demonstrated competitive—and, in some cases, superior—performance compared to GNNs across a variety of graph mining tasks~\cite{nguyen2020dataset, wang2024fast}.

\begin{figure}[]
  \centering
  \includegraphics[width=0.6\linewidth]{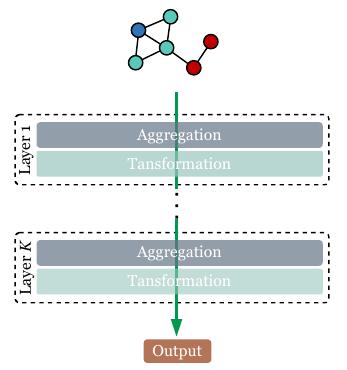}
  \caption{The layer stacking structure.}\label{fig:stack}
\end{figure}

In spite of its demonstrated effectiveness, the GNTK exhibits certain inherent limitations. Specifically, the GNTK employs a layer-stacking strategy to compute the kernel value, as illustrated in Figure~\ref{fig:stack}. In this strategy, each layer begins with an aggregation step, where information from neighboring nodes is collected, followed by a transformation step that applies non-linear operations to the aggregated data~\cite{jacot2018neural}. These steps are performed recursively across layers to iteratively update the kernel values. While this recursive stacking mechanism enhances the expressiveness of the kernel by incorporating information from multiple layers, it also leads to an increase in computational complexity, which grows linearly with the number of layers stacked.

Moreover, as the depth of the network increases, the transformations performed in the deeper layers contribute progressively less to the overall performance of the model~\cite{wu2019simplifying}. This diminishing contribution limits the utility of excessively deep networks. As a result, while the GNTK is a powerful tool, its high computational cost and reduced benefits at greater depths may hinder its scalability and applicability in certain cases.


In this paper, we investigate the structure of the GNTK method and identify computational redundancies. Specifically, as the number of stacked layers increases, the impact of the non-linear transformations in deeper layers diminishes. Thus, retaining only aggregations and a limited number of Neural Tangent Kernel~\cite{jacot2018neural} iterations within these layers suffices. This approach reduces computation time while maintaining the kernel function's effectiveness.

Building on this observation, we propose the \textbf{S}implified \textbf{G}raph Neural \textbf{T}angent \textbf{K}ernel (\textbf{SGTK}) method. This method involves performing $K$ consecutive steps of aggregation operations, followed by a single update of the kernel function. SGTK reduces computational redundancy and enhances the efficiency of kernel function computation.
Furthermore, we define an infinitely wide, simple graph neural model, deriving a Gaussian process (GP)~\cite{williams2006gaussian} kernel function with reduced computational complexity, termed the \textbf{S}implified \textbf{G}raph \textbf{N}eural \textbf{K}ernel (\textbf{SGNK}). This method calculates the expectations of the post-activation output of the infinitely wide graph model to determine the kernel value. Compared to SGTK, SGNK further streamlines the process and reduces computational complexity while maintaining effectiveness in downstream tasks.
Our main contributions are as follows:
\begin{itemize}
    \item 
    We propose SGTK, which replaces the stacking layers in GNTK with continuous $K$-step message passing to reduce computational redundancy.
    \item By interpreting the infinite-width simple graph neural model as a GP, we introduce SGNK, which further reduces computational complexity compared to SGTK.
    \item
    We conducted a theoretical analysis to demonstrate the low computational complexity of the proposed methods and validated their computational efficiency improvements through extensive experiments.
    \item 
    We have experimentally validated the proposed methods in both node and graph classification tasks, demonstrating that our methods ensure excellent classification accuracy.

\end{itemize}

\section{Related Works}
\label{sec:relatedworks}

\noindent \textbf{Graph Neural Networks}. GNNs have undergone significant advancements, addressing diverse aspects of graph learning through various models. For semi-supervised learning, GCNs~\cite{kipf2016semi} leverage spectral graph convolutions, while GATs~\cite{velickovic2017graph} incorporate attention mechanisms to prioritize relevant neighbors. Extensions like R-GCNs~\cite{schlichtkrull2018modeling} handle multi-relational data, and GraphSAGE~\cite{hamilton2017inductive} introduces sampling-based aggregation for scalable inductive learning. To enhance expressiveness, GINs~\cite{xu2018powerful} employ sum aggregation inspired by the Weisfeiler-Lehman test. Simplifying computations, SGC~\cite{wu2019simplifying} removes non-linear activations and collapses weight matrices, while APPNP~\cite{gasteiger2018predict} integrates personalized PageRank with neural networks for improved performance.

\noindent \textbf{Graph Neural Tangent Kernel}. 
Neural tangent kernel (NTK)~\cite{jacot2018neural} fundamentally transforms our understanding of training dynamics in infinitely wide neural networks by equating them to kernel methods. 
CNTK~\cite{arora2019exact} extends NTK to convolutional neural networks. GNTK~\cite{du2019graph} and GPGC~\cite{hu2020infinitely} are introduced for graph neural networks, demonstrating their efficacy in graph-based tasks. 
Krishnagopal et al.~\cite{krishnagopal2023graph} investigate the training dynamics of large-graph GNNs using GNTKs and graphons. GCNTK~\cite{zhou2023explainability} shows that for wide GCNs, the output for semi-supervised problems can be described by a simple differential equation. A quantum graph learning model, GraphQNTK~\cite{tang2022graphqntk}, leverages quantum parallelism to enhance computational efficiency; however, experiments reveal that its performance lags behind that of GNTK. 
Jiang et al.~\cite{jiang2022fast} employ sketching to accelerate GNTK, but they do not experimentally demonstrate efficiency improvements. 
Wang et al.~\cite{wang2024fast} integrate structural information with the neural tangent kernel and propose the SNTK, a method designed to address graph condensation problems. Additionally, GCKM~\cite{wu2024graph} employs stacked layers to compute graph kernels, but this approach involves redundant computations, which contribute to increased computational complexity.

\section{Preliminary}
In this section, we first present the GPs. Then, the dynamical equation of infinitely wide neural networks during training and the NTK \cite{jacot2018neural} are introduced, which converge to a deterministic kernel and remain unchanged throughout the training process.

\subsection{Gaussian Process}
GPs~\cite{neal2012bayesian} are Bayesian approaches for modeling and inferring unknown functions by leveraging correlations between data points. Specifically, they employ a kernel (or covariance) function to quantify the similarity between data points. GPs allow for predictions on new input data while providing a measure of uncertainty in these predictions based on the observed data. Consequently, GPs can flexibly capture complex patterns in the data without requiring an explicit specification of the functional form~\cite{williams2006gaussian}.

A straightforward example of a GP can be derived from a Bayesian linear regression model defined as $f({x}) = {W}^{\top}\phi({x})$, where the weights follow a prior distribution ${W}\sim \mathcal{N}({0},{\sigma}^2_w)$. In practical scenarios, we often observe noisy observations $y = f(x) + b$, where $b$ is additive Gaussian noise, distributed as $b \sim \mathcal{N}(0,\sigma^2_{b})$.
Given the observed data $\mathcal{X}$ and the kernel (or covariance) function $\mathcal{K}$, the joint prior distribution of the training outputs $\mathcal{Y}$ and the test outputs $f_t$ under the prior can be expressed as:
\begin{align}
    \begin{bmatrix}
    \mathcal{Y}\\
    f_t
    \end{bmatrix} \sim \mathcal{N} \left (\bs{0},\begin{bmatrix}
    \mathcal{K}(\mathcal{X}, \mathcal{X})+\sigma^2_{b}\bs{I} &\mathcal{K}(\mathcal{X}, {x}_t)  \\
    \mathcal{K}({x}_t, \mathcal{X}) &\mathcal{K}({x}_t, {x}_t) \end{bmatrix}\right ).
\end{align}
The predictive equation for GP regression is
\begin{align}
    f_{t}|{x}_{t}, \mathcal{X}, \mathcal{Y} \sim \mathcal{N}( {\mu}, Cov(f_t)),
\end{align}
where,
\begin{align}
    {\mu}    & = \mathcal{K}({x}_t,\mathcal{X}) [\mathcal{K}(\mathcal{X},\mathcal{X})+\sigma^2_{b}\bs{I}]^{-1}\mathcal{Y},  \\
    Cov(f_t) & = \notag \\  
    \mathcal{K}({x}_t,{x}_t) & - \mathcal{K}({x}_t, \mathcal{X})[\mathcal{K}(\mathcal{X},\mathcal{X})+\sigma^2_{b}\bs{I}]^{-1}\mathcal{K}(\mathcal{X},{x}_t).
\end{align}

\subsection{Neural Tangent Kernel}\label{GNTK}
Neural Tangent Kernel (NTK) is a kernel method derived from infinite-width neural networks, used to understand and analyze the training dynamics of neural networks~\cite{jacot2018neural, hu2019simple, lee2019wide}.
Given the regularized mean-squared loss of the infinite-width neural network model $f_{\theta}$, formulated as follows: 
\begin{align}\label{eq:loss}
\begin{gathered}
    \mathcal{L} = \frac{1}{2}|| f_{\theta}(\mathcal{X}) - \mathcal{Y} ||^{2}_{F} + \frac{\lambda}{2}||\theta||^{2}_{F},
\end{gathered}
\end{align}
where $\lambda > 0$ is the regularization parameter, and $\mathcal{X}$ and $\mathcal{Y}$ represent the training data and the corresponding labels, respectively.
Let $\phi(x) = \nabla_{\theta}f_{\theta}(x)$.
Using the Stochastic Gradient Descent (SGD) algorithm, the optimization of the model parameter $\theta$ can be expressed as an ordinary differential dynamic equation when the learning rate is sufficiently small~\cite{lee2019wide}:
\begin{equation}\label{eq:dyna}
\begin{aligned}
    \dot{\theta}  & = - \nabla_{\theta}\mathcal{L} = - \frac{\partial f_{\theta}(\mathcal{X})}{\partial \theta} \cdot \frac{\partial \mathcal{L} }{\partial f_{\theta}} \\
    & = -\phi(\mathcal{X})(f_{\theta}(\mathcal{X})-\mathcal{Y}) - \lambda(\theta) \\
    & = -\phi(\mathcal{X})\phi(\mathcal{X})^{T}\theta + \phi(\mathcal{X})\mathcal{Y} - \lambda \theta,
    \end{aligned}
\end{equation}
where $\mathcal{K} = \phi(\mathcal{X})^{T}\phi(\mathcal{X})$ is the {Neural Tangent Kernel}. 
\begin{figure*}[!htbp]
  \centering
  \includegraphics[width=0.85\linewidth]{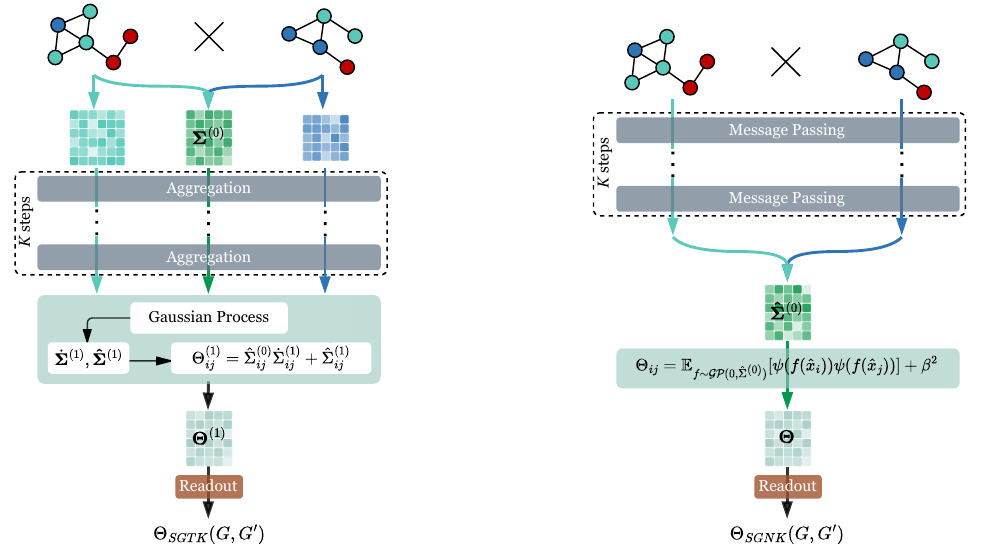}
  \caption{This kernel structure includes SGTK and SGNK.
The SGTK improves computational efficiency by using continuous $K$-step aggregation. Furthermore, the SGNK models the infinite-width simple graph neural network as a GP, requiring only the expectation of the infinite-width neural model's post-activation output to determine the kernel values. This approach further reduces computational complexity.}
\label{fig:sgnks}
\end{figure*}

\section{Methodology}
\label{sec:methodlogy}
\newcommand{\Ein}{{\vE_{\text{in}}}}


In this section, we begin with a brief review of the GNTK to establish the necessary background. Thereafter, we introduce the proposed SGTK and SGNK, as illustrated in Figure~\ref{fig:sgnks}. Subsequently, we present the implementation details of the proposed methods, outlining the key steps and considerations. Lastly, we provide a theoretical analysis of computational complexity, showing that the proposed kernel methods are more efficient than previous GNTK approaches.

\subsection{Notations and Definitions} 

In general, a graph can be represented as $G = (\mathcal{V},\mathcal{E})$, where $\mathcal{V} = \{v_1, v_2, \cdots, v_{|\mathcal{V}|}\}$ denotes the set of $|\mathcal{V}|$ nodes, and $\mathcal{E}$ represents the edge set. We use $X = [x_1, x_2, \cdots, x_{|\mathcal{V}|}]^{\top} \in \mathbb{R}^{|\mathcal{V}| \times d}$ to denote the node feature matrix, where $d$ is the dimensional size of the node features. The structural information of the graph is captured by the adjacency matrix $A \in \{0,1\}^{|\mathcal{V}| \times |\mathcal{V}|}$, where $A_{ij} = 1$ indicates a connection between nodes $v_i$ and $v_j$, and $A_{ij} = 0$ otherwise.
In this study, we examine both graph-level and node-level downstream tasks. Given two graphs $G_1 = \{X_1,A_1\}$ and $G_2 = \{X_2, A_2\}$ with $n$ and $m$ nodes, respectively, the graph kernel is defined as $\mathcal{K}(G_1,G_2):\mathbb{R}^{n \times d} \times \mathbb{R}^{m \times d} \to \mathbb{R}$. Conversely, the node-level kernel is defined as $\mathcal{K}(v_i,v_j):\mathbb{R}^{d} \times \mathbb{R}^{d} \to \mathbb{R}$.

\subsection{Brief Review of Graph Neural Tangent Kernel}
The GNTK~\cite{du2019graph} is a kernel function derived from infinite-width graph neural networks. It serves as a tool for analyzing and understanding the training behavior of GNNs. Specifically, we consider a two-layer GCN~\cite{kipf2016semi}, which is defined as follows:
\begin{align}\label{eq:GCN}
        f_{\mathrm{GCN}}(X,A) =
        \mathrm{softmax}\{ \hat{A} \ \mathrm{ReLU}\{ \hat{A} X W^{(0)} \} W^{(1)} \},
\end{align}
where $\hat{A} = \tilde{D}^{-\frac{1}{2}}\tilde{A} \tilde{D}^{-\frac{1}{2}}$ is the spectral filter, $\tilde{A} = A + I$, and $\tilde{D} = \mathrm{diag}(\sum_{j}{\hat{A}_{1j}}, \sum_{j}{\hat{A}_{2j}}, \ldots, \sum_{j}{\hat{A}_{nj}})$.
The GNTK follows a structure similar to that of the GCN. It first uses adjacency information to aggregate the covariance matrix of the node features. This matrix is then input into the NTK iterations. Repeating these steps $K$ times yields the kernel value for an infinite-width GCN with $K$ layers, as illustrated on the left in Figure~\ref{fig:sgnks}.
For two graphs $G$ and $G'$, the GNTK kernel matrix is computed as follows:
\begin{align}\label{eq:GNTK}
        \bs{\Theta} = \mathrm{NTK}\{ \hat{A}_1 \  \mathrm{NTK}\{ \hat{A}_1 \boldsymbol{\Sigma}^{(0)} \hat{A}_2^{\top} \} \hat{A}_2^{\top} \},
\end{align}
where $\Sigma^{(0)}_{ij} = c_i c_j x_i \cdot x_j$, and $c_{i} = ( \| \sum_{p \in \mathcal{N}(i) \cup \{i\}} x_{p} \|_{2} )^{-1}$ serves as a scaling coefficient that facilitates subsequent NTK computations. $x_i \in G_1$ and $x_j \in G_2$.
By conducting a readout operation on the kernel matrix in Equation~\eqref{eq:GNTK}, the similarity value between graphs $G_1$ and $G_2$ can be derived.

The performance and efficiency of GNTK greatly depend on its architecture, including the number of layers and NTK iterations.
Although iterative stacking of layers enables GNTKs to match GCNs of arbitrary depths, minimally contributing NTK iterations in deeper layers introduce computational redundancy.
This results in computational overhead, limiting their practical applicability~\cite{wu2019simplifying, du2019graph}.

\subsection{Simplified Graph Neural Tangent Kernel}
GNTK originates from the infinitely wide GCN. To simplify the redundant computations in GNTK, it is essential to first examine the computational complexity inherent in the infinitely wide GCN.

We propose streamlining the GCN structure by retaining $K$ rounds of message passing to enhance the receptive field of node representations while reducing the nonlinear transformations to collapse the weight matrix.
Specifically, we consider the infinite-width graph neural network model with $K$-step message passing.
\begin{align}\label{eq:gc}
\hat{X} = \hat{A}^{K}{X},
\end{align}
where $\hat{X} \in \mathbb{R}^{n \times d}$, with $n$ representing the number of samples and $d$ the input dimensionality. 

To enhance the receptive field of the node's features, a two-layer neural network is employed, where the width of the hidden layer approaches infinity.
Let the infinite-width neural network take $\hat{X}$ as its input.
Each layer consists of a biased linear transformation $f^{(l)}: \mathbb{R}^{n \times d^{(l-1)}} \to \mathbb{R}^{n \times d^{(l)}}$, where $d^{(l)}$ is the width of the $l$-th layer.
The parameters are initialized as $W^{(l)} \sim \mathcal{N}(0, \sigma_w^2)$ and $b^{(l)} \sim \mathcal{N}(0, \sigma_b^2)$. Let $\hat{X}^{(0)} = \hat{X}$. The output of the $l$-th layer of the network can be expressed as follows:
\begin{align}\label{eq:mlp}
    f^{(l)}(\hat{X}^{(l)}) &= \sigma_w \hat{X}^{(l)}W^{(l)} + \beta b^{(l)}, \\ 
    \hat{X}^{(l+1)} &= \psi(f^{(l)}(\hat{X}^{(l)})), 
\end{align}
where $\beta$ is a hyperparameter adjusting the effect of the bias on training. Let $\sigma^2_b = 1$. Under LeCun initialization~\cite{lecun2002efficient}, the variance of the weights $W^{(l)}$ is denoted as $\sigma_w^2 = \frac{1}{d^{(l)}}$.
As the width of a neural network approaches infinity, its training dynamics can be represented by an explicit limiting kernel~\cite{jacot2018neural}, which we term the Simplified Graph Neural Tangent Kernel (SGTK).

Specifically, for two graphs \( G_1 \) and \( G_2 \), with node features \( X_1 = \{x_i\}_{i=1}^n \) and \( X_2 = \{x_j\}_{j=1}^{m} \), respectively, the kernel preprocessing involves the covariance matrix of the enhanced node features:
\begin{align}
      \hat{\Sigma}^{(0)}_{ij} = \frac{1}{d} \hat{x_i} \hat{x_j}^{\top} + \beta^{2}.  \label{eq:init}
\end{align}

Suppose the enhanced node features $\{\hat{x}_i\}_{i=1}^{n}$ and $\{\hat{x}_j\}_{j=1}^{m}$ are independently and identically distributed (i.i.d.). 
In the limit as the layer width tends to infinity, the node-level SGTK kernel value between node $v_i$ of $G_1$ and node $v_j$ of $G_2$ is defined by:
\begin{align}
    {\Theta}^{(1)}_{ij} &= \hat{\boldsymbol{\Sigma}}^{(0)}_{ij}  \dot{\boldsymbol{\Sigma}}^{(1)}_{ij} + \hat{\boldsymbol{{\Sigma}}}^{(1)}_{ij}, \label{eq:sgtk}
\end{align}
where the intermediate calculation variables of the covariance matrix are defined as:
\begin{align}
    \boldsymbol{\hat{\Sigma}}^{(1)}_{ij} =&\mathbb{E}_{f^{(0)} \sim \mathcal{GP}(0,\hat{\bs{\Sigma}}^{(0)})}   [\psi(f^{(0)}(\hat{x}_i))\psi(f^{(0)}(\hat{x}_j))] + \beta^{2}, \label{eq:sigma_hat} \\
    \dot{\boldsymbol{\Sigma}}^{(1)}_{ij} =&\mathbb{E}_{f^{(0)} \sim \mathcal{GP}(0,\hat{\bs{\Sigma}}^{(0)})}   [\dot{\psi}(f^{(0)}(\hat{x}_i))\dot{\psi}(f^{(0)}(\hat{x}_j))], \label{eq:sigma_dot}
\end{align}
where $\dot{\psi}(\cdot)$ is the derivative of the activation function.
As illustrated in Figure~\ref{fig:sgnks}, compared to GNTK, which requires iterative stacking of message passing and NTK iterations, the proposed SGTK enhances the receptive field of node features through $K$-step aggregation in a concentrated manner. This is followed by a single kernel update, effectively collapsing the layers to reduce computational redundancy.

\subsection{Simplified Graph Neural Kernel}
SGTK streamlines the computation process of GNTK by reducing redundant calculations. However, the computation of SGTK relies on the characteristics of infinite-width neural networks and the Gaussian process kernel.
The computational efficiency of the kernel method is limited by its reliance on the Gaussian process kernel.

We further propose a kernel method with lower computational complexity than SGTK, termed the Simplified Graph Neural Kernel (SGNK). This approach directly models the defined infinite-width simple graph neural network as a Gaussian process, which is determined by the expectations of the post-activation outputs of the Gaussian process~\cite{neal2012bayesian,lee2017deep}. 
These expectations can be computed analytically without the need for iteration. Consequently, this direct kernel method further reduces computational complexity. 
For the node features $x_i$ and $x_j$, the Gaussian process kernel is defined recursively by the following functions:
\begin{align}
    \Sigma^{(0)}(x_i,x_j) &= \frac{1}{d} x_i x_j^{\top} + \beta^{2}, \\
    \Sigma^{(l+1)}(x_i,x_j) &= 
    \mathbb{E}_{f^{(l)} \sim \mathcal{GP}(0,\boldsymbol{\Sigma}^{(l)})} 
    \left[\psi(f^{(l)}(x_i))\psi(f^{(l)}(x_j))\right] + \beta^{2}.
\end{align}

For an infinitely wide simple graph neural network with $K$-step message passing, followed by a two-layer neural network, the node-level Simplified Graph Neural Kernel (SGNK) for the node features ${x}_i$ and ${x}_j$ is defined as follows:
\begin{align}
    {\Theta}_{ij} =  \mathbb{E}_{f^{(0)} \sim \mathcal{GP}(0,\hat{\bs{\Sigma}}^{(0)})}   [\psi(f^{(0)}(\hat{x}_i))\psi(f^{(0)}(\hat{x}_j))] + \beta^{2}. \label{eq:gp}
\end{align}

Let the neural network adopt the Error Function as the activation function, defined as $ \psi(x) = \frac{2}{\pi}\int_{0}^{x} e^{-t^2} dt $, then the SGNK can be calculated analytically~\cite{williams1996computing},
\begin{align}
    {\Theta}_{ij} =  \frac{2}{\pi} sin^{-1}\frac{2\tilde{x}^{\top}_i\Sigma_{w} \tilde{x}_j }{\sqrt{(1+2\tilde{x}^{\top}_i\Sigma_w \tilde{x}_i)(1+2\tilde{x}^{\top}_j\Sigma_w \tilde{x}_j)} }, \label{eq:sgnk}
\end{align}
where $\tilde{x} = (\hat{x}, 1)^{\top}$ is the augmented vector, and the variance matrix $\Sigma_w = \text{diag}(1, 1, \cdots, \sigma^2_b)$.
In matrix form, the calculation process can be expressed as follows:
\begin{align}
    \bs{\Theta} = \frac{2}{\pi} sin^{-1}\frac{2\tilde{X}_1\Sigma_w\tilde{X}_2^{\top}}{\sqrt{[\bs{1} +2diag(\tilde{X}_1\Sigma_w\tilde{X}_1^{\top})]^{\top}[\bs{1} +2diag(\tilde{X}_2\Sigma_w\tilde{X}_2^{\top})]}}.
\end{align}

Given that $\Sigma_w = diag(1, 1, \cdots, \sigma^2_b)$, we have, 
\begin{align}
    \bs{\Theta} = \frac{2}{\pi} sin^{-1}\frac{2\tilde{X}_1\tilde{X}_2^{\top}}{\sqrt{[\bs{1} +2diag(\tilde{X}_1\tilde{X}_1^{\top})]^{\top}[\bs{1} +2diag(\tilde{X}_2\tilde{X}_2^{\top})]}} \label{eq:matrix}
\end{align}

Finally, for the graph-level tasks, the readout operation yields the kernel value between $G$ and $G'$, which is:
\begin{align}
    \Theta(G,G') =  \sum_{i\in G, j \in G'}\Theta_{ij}. \label{eq:readout}
\end{align}
The detailed execution processes of our proposed SGTK and SGNK are shown in the Algorithm~\ref{alg1} and ~\ref{alg2}.

\begin{algorithm}[t]
	\caption{Simplified Graph Neural Tangent Kernel (SGTK)}
    \label{alg1}
	\begin{algorithmic}[1]
        \STATE \textbf{Input:} Graphs $G= \{X,A\}$, $G' = \{X',A' \}$.\\
        \STATE \textbf{Output:} Node-level kernel matrix $\bs{\Theta}^{(1)}$ and graph-level kernel value $\Theta(G,G')$.\\
        \STATE \textbf{Initialize:} $\boldsymbol{\Sigma}^{(0)} = X(X')^{\top}$, $\boldsymbol{\Sigma}^{(0)}_{X} = X(X)^{\top}$, $\boldsymbol{\Sigma}^{(0)}_{X'} = X'(X')^{\top}$, $\hat{A} = D^{-\frac{1}{2}} ( A+I ) D^{-\frac{1}{2}}$, $\hat{A}' = D^{-\frac{1}{2}} ( A'+I ) D^{-\frac{1}{2}}$, $K$. \\
        \STATE $K$-step aggregation on covariance matrix.
        \begin{align}
            \hat{\boldsymbol{\Sigma}}^{(0)} &= \frac{1}{d}\hat{A}^{K} \boldsymbol{\Sigma}^{(0)} (\hat{A'}^{K})^{\top} + \beta^2, \notag \\
            \hat{\boldsymbol{\Sigma}}_{X}^{(0)} &= \frac{1}{d}\hat{A}^{K} \boldsymbol{\Sigma}_{X}^{(0)} (\hat{A}^{K})^{\top}+ \beta^2, \notag \\
            \hat{\boldsymbol{\Sigma}}_{X'}^{(0)} &= \frac{1}{d}\hat{A'}^{K} \boldsymbol{\Sigma}_{X'}^{(0)} (\hat{A'}^{K})^{\top}+ \beta^2. \notag
        \end{align} \\
        \STATE  Update $\hat{\bs{\Sigma}}_{X}^{(1)}$ and $\hat{\bs{\Sigma}}_{X'}^{(1)}$ by Eq.~\eqref{eq:sigma_hat}. \\
        \STATE  Update $\hat{\bs{\Sigma}}^{(1)}$ and $\dot{\bs{\Sigma}}^{(1)}$ by Eq.~\eqref{eq:sigma_hat}, \eqref{eq:sigma_dot}, $\hat{\bs{\Sigma}}^{(0)}$, $\hat{\bs{\Sigma}}_{X}^{(1)}$, and $\hat{\bs{\Sigma}}_{X'}^{(1)}$. \\
        \STATE Calculate the nodel-level kernel value by Eq.~\eqref{eq:sgtk},
        \begin{align}
        {\Theta}^{(1)}_{ij} &= \hat{\boldsymbol{\Sigma}}^{(0)}_{ij}  \dot{\boldsymbol{\Sigma}}^{(1)}_{ij} + \hat{\boldsymbol{{\Sigma}}}^{(1)}_{ij}. \notag
        \end{align}  \\
        \STATE Readout, $\Theta(G,G') = \sum_{ij} {\Theta}^{(1)}_{ij}$.
	\end{algorithmic}
\end{algorithm}

\begin{algorithm}[t]
	\caption{Simplified Graph Neural Kernel (SGNK)}
    \label{alg2}
	\begin{algorithmic}[1]
        \STATE \textbf{Input:} Graphs $G= \{X,A\}$, $G' = \{X',A' \}$.\\
        \STATE \textbf{Output:} Node-level kernel matrix $\bs{\Theta}$ and graph-level kernel value $\Theta(G,G')$.\\
        \STATE \textbf{Initialize:} $\hat{A} = D^{-\frac{1}{2}} ( A+I ) D^{-\frac{1}{2}}$, $\hat{A}' = D^{-\frac{1}{2}} ( A'+I ) D^{-\frac{1}{2}}$, $K$. \\
        \STATE $K$-step message passing on nodes features.
        \begin{align}
            \hat{X} = \hat{A}^{K} X, \quad \hat{X}'= \hat{A}'^{K} {X'}, \notag
        \end{align} \\
        \STATE Calculate the node-level kernel matrix by Eq.~\eqref{eq:matrix},
        \begin{align}
            \bs{\Theta} = \frac{2}{\pi} sin^{-1}\frac{2\tilde{X}_1\tilde{X}_2^{\top}}{\sqrt{[\bs{1} +2diag(\tilde{X}_1\tilde{X}_1^{\top})]^{\top}[\bs{1} +2diag(\tilde{X}_2\tilde{X}_2^{\top})]}} 
        \end{align} \\
        \STATE Readout, $\Theta(G,G') = \sum_{ij} {\Theta}_{ij}$.
	\end{algorithmic}
\end{algorithm}

\subsection{Computational Complexity Analysis}\label{complex}
To demonstrate the reduced computational complexity of our proposed simplified methods compared to the GNTK, we present a comprehensive computational complexity analysis. The notations used are as follows: $n$ and $m$ represent the number of nodes in two different graphs, $d$ stands for the node feature dimensionality, $k$ indicates the average degree per node, and $K$ represents the number of message-passing iterations.

In each layer, the computational complexity of GNTK comprises two parts: message passing and NTK iterations. The complexity of message passing is $O(nk^{2})$, while the NTK iterations have a complexity of $O(n^2d + nmd + m^2d)$. Consequently, for the GNTK with $K$ layers, the total computational complexity is $O(K(nk^{2} + n^2d + m^2d + nmd))$.

In contrast, the proposed SGTK performs $K$ steps of aggregation and eliminates multi-layer stacking, achieving an overall complexity of $O(Knk^{2} + n^2d + m^2d + nmd)$. Furthermore, the SGNK reduces this complexity to $O(Knk^{2} + nmd)$. This indicates that the complexity of our proposed kernels is lower than that of GNTK.
\section{Experiment}
\label{sec:Experiments}

\begin{table*}[h]
\caption{Graph-level dataset details.}
\label{tab:graph_dataset}
\centering
\begin{tabular}{crrrrr}
\hline
Dataset     & \multicolumn{1}{c}{\# Graphs} & \multicolumn{1}{c}{Avg. Nodes} & \multicolumn{1}{c}{Avg. Edges} & \multicolumn{1}{c}{\# Class} & \multicolumn{1}{c}{\# Node feature} \\ \hline
IMDB-BINARY & 1,000                         & 19.8                           & 96.5                           & 2                            & -                                   \\
IMDB-MULTI  & 1,500                         & 13                             & 65.9                           & 3                            & -                                   \\
PTC\_MR     & 344                           & 14.3                           & 14.7                           & 2                            & -                                   \\
MUTAG       & 188                           & 17.9                           & 19.8                           & 2                            & -                                   \\ \hline
\end{tabular}
\end{table*}
\begin{table}[h]
\caption{Node-level dataset details.}
\label{tab:node_dataset}
\centering
\begin{tabular}{crrrr}
\hline
Dataset   & \multicolumn{1}{c}{\# Nodes} & \multicolumn{1}{c}{\# Edges}   & \multicolumn{1}{c}{\# Class} & \multicolumn{1}{c}{\# Node feature} \\ \hline
Cora      & 2,708                        & 5,429                          & 7                            & 1,433                               \\
Citeseer  & 3,327                        & 9,104                          & 6                            & 3,703                               \\
Pubmed    & 19,717                       & 44,338                         & 3                            & 500                                 \\
Photo     & 7,650                        & {\color[HTML]{404040} 238,162} & 8                            & 745                                 \\
Computers & 13,752                       & 491,722                        & 10                           & 767                                 \\ \hline
\end{tabular}
\end{table}
This section evaluates the effectiveness of the proposed kernel methods through extensive experiments on both graph and node classification tasks. We compare the performance and efficiency of our methods with neural network approaches and other graph kernel methods.
Our experimental findings demonstrate the robust performance of our approach in classification tasks. Furthermore, they highlight computational efficiency improvements compared to neural networks and GNTK. The experiments were conducted on an Apple M1 Pro CPU with 16GB of RAM.
\subsection{\textbf{Experimental Settings}}



\noindent \textbf{Datasets}.
Five widely-used node-level benchmark datasets are selected to evaluate the effectiveness of our methods: Cora, CiteSeer, PubMed~\cite{yang2016revisiting}, Photo, and Computers~\cite{shchur2018pitfalls}.
Additionally, four commonly used graph-level datasets are selected to evaluate the graph similarity modeling capability of the proposed kernel method: IMDB-BINARY, IMDB-MULTI~\cite{pinar2015deep}, PTC\_MR~\cite{Nils2012Subgraph} and MUTAG~\cite{kriege2012subgraph}. The details of these datasets are presented in Table~\ref{tab:node_dataset} and Table~\ref{tab:graph_dataset}.
For node classifications, we utilize publicly available splits for the Cora, CiteSeer, and PubMed datasets. For the Computers dataset, the first 20 instances from each category are selected as the training set, while the last 100 instances of each class are chosen as the testing set. For graph datasets lacking explicit node features, node degrees are converted into one-hot encoded representations. 

\noindent  \textbf{Baselines}.
To evaluate the effectiveness of the proposed SGTK and SGNK, we compare them against various SOTA graph neural networks and kernel methods. The baseline graph neural networks include GCN~\cite{kipf2016semi}, GAT~\cite{velickovic2017graph}, SGC~\cite{wu2019simplifying}, GraphSAGE~\cite{hamilton2017inductive}, APPNP~\cite{gasteiger2018predict}, and GIN~\cite{xu2018powerful}. Additionally, we consider graph kernels such as the WL subtree~\cite{shervashidze2011weisfeiler}, RetGK~\cite{zhang2018retgk}, GraphQNTK~\cite{tang2022graphqntk}, GNTK~\cite{du2019graph}, SNTK~\cite{wang2024fast}, and GCKM~\cite{wu2024graph}.

\noindent \textbf{Parameter Settings}. 
We tune the hyperparameter $K$ over the range $\{1, 2, 3, 4, 5\}$ for GCKM, SNTK, GNTK, SGTK, and SGNK. For classification using SVM, its hyperparameter is optimized over the range $[10^{-2}, 10^{4}]$. Similarly, for KRR, the regularization parameter $\lambda$ is tuned over the range $[10^{-2}, 10^{2}]$.

\subsection{\textbf{Node Classification Performance Comparison}}

This subsection explores the application of kernel functions to node classification tasks. Specifically, we apply the proposed kernel methods to node classification datasets, utilizing an SVM classifier to evaluate their performance. For GNN methods, we report the mean and standard deviation of the results from 10 runs and present the classification accuracy of the kernel methods using fixed training and testing set splits.

\begin{table}[]
\caption{Node classification accuracy (\%). The top three performance methods are highlighted in varying shades of green, with darker shades representing superior performance.}
\label{tab:node_acc}
\centering
\scalebox{0.99}{
\setlength{\tabcolsep}{3pt} 
\begin{tabular}{cccccc}
\hline
Methods             & Cora                             & Citeseer                      & Pubmed                        & Photo                            & Computers                        \\ \hline
GCN                 & \cellcolor[HTML]{EBF1DE}81.1±0.5 & \cellcolor[HTML]{EBF1DE}71.7±0.1                      & 77.1±0.3                      & \cellcolor[HTML]{EBF1DE}91.4±0.8 & 84.2±0.3                         \\
GAT                 & 79.5±1.0                         & 68.8±0.2                      & 75.7±0.9                      & 90.1±0.6                         & 83.3±0.7                         \\
SGC                 & 79.0±0.3                         & 67.4±0.2                      & 76.9±0.1                      & 91.0±0.5                         & 84.1±0.2                         \\
APPNP               & 79.3±0.2                         & 67.1±1.0                      & 76.0±0.5                      & \cellcolor[HTML]{D8E4BC}91.7±0.3 & \cellcolor[HTML]{D8E4BC}86.3±0.2 \\
GraphSAGE           & 79.3±0.4                         & 67.5±0.3                      & 73.9±0.6                      & 90.0±0.1                         & 82.8±0.6                         \\
GIN                 & 77.2±1.8                         & 67.4±0.2                      & 78.1±0.1                      & 43.8±5.5                         & 30.0±5.2                         \\
GCKM                & 82.4                             & \cellcolor[HTML]{D8E4BC}72.3  & \cellcolor[HTML]{EBF1DE}79.8  & 91.30                            & 84.20                            \\
SNTK                & 80.70                            & 69.20                         & 78.40                         & 91.00                            & 84.70                            \\
GNTK                & 80.50                            & 69.20                         & 77.20                         & 91.00                            & 84.70                            \\
\textbf{SGTK(Our)} & \cellcolor[HTML]{D8E4BC}82.60    & 71.10 & \cellcolor[HTML]{C4D79B}80.00 & \cellcolor[HTML]{C4D79B}91.75    & \cellcolor[HTML]{C4D79B}86.50    \\
\textbf{SGNK(Our)} & \cellcolor[HTML]{C4D79B}82.80    & \cellcolor[HTML]{C4D79B}72.60 & \cellcolor[HTML]{D8E4BC}79.90 & 90.88                            & \cellcolor[HTML]{EBF1DE}85.50    \\ \hline
\end{tabular}
}
\end{table}

As shown in Table~\ref{tab:node_acc}, the proposed methods, SGTK and SGNK, consistently demonstrate strong performance across all five node classification datasets. These methods achieve the highest classification accuracy in each dataset, outperforming both traditional kernel methods and GNNs. This consistent superiority underscores the effectiveness of the proposed kernel methods in capturing complex graph structures and node relationships.

The results obtained from the node classification datasets further emphasize the effectiveness of the proposed methods in addressing node classification tasks. Specifically, the capability of SGTK and SGNK to maintain high accuracy across diverse datasets highlights their robustness and generalizability. These findings suggest that the proposed methods are well-suited for graph-based learning tasks, contributing to advances in graph analysis techniques.


To comprehensively evaluate the classification performance of the proposed kernel methods, experiments were also conducted using Kernel Ridge Regression (KRR)~\cite{vovk2013kernel}, as detailed in Table~\ref{tab:node_krr}. KRR, a widely used technique for regression tasks, ensures reliable performance and demonstrates the compatibility of the proposed methods with various classifiers. Given the set of training nodes $\mathcal{V}$ and their corresponding labels $\mathcal{Y}$, KRR models the optimal predictions for test nodes $v_t$ by leveraging the kernel function to capture relationships between nodes. Furthermore, the proposed methods are compatible with Support Vector Machines (SVMs), showcasing their adaptability across different classification frameworks. The prediction for a test node $v_t$ is computed as follows:
\begin{equation}
f(x_t) = \mathcal{K}(v_t, \mathcal{V})[\mathcal{K}(\mathcal{V}, \mathcal{V}) + \lambda I]^{-1} \mathcal{Y},
\end{equation}
where the kernel function $\mathcal{K}(v, v') : \mathbb{R}^d \times \mathbb{R}^d \to \mathbb{R}$ quantifies the similarity between nodes $v$ and $v'$, and $\lambda > 0$ is the regularization parameter that controls the trade-off between fitting the training data and maintaining model generalizability.
By incorporating the kernel function $\mathcal{K}$, KRR effectively captures the complex relationships among nodes in the graph, making it a suitable choice for evaluating the proposed kernel methods. 


The experimental results, summarizing the performance of the kernel methods under the KRR framework, are presented in Table~\ref{tab:node_krr}. These results provide a comprehensive comparison and highlight the advantages of the proposed methods over existing approaches.

\begin{table}[h]
\caption{Kernel method on kernel ridge regression results.}
\label{tab:node_krr}
\centering
\setlength{\tabcolsep}{3pt} 
\begin{tabular}{cccccc}
\hline
Methods & Cora                          & Citeseer                      & Pubmed                        & Photo                         & Computers                     \\ \hline
GCKM    & 82.50                         & 71.70                         & 79.90                         & \cellcolor[HTML]{C4D79B}93.25 & 86.80                         \\
SNTK    & 81.40                         & 70.70                         & 80.00                         & 90.13                         & 84.20                         \\
GNTK    & 81.30                         & 71.10                         & 78.60                         & 89.63                         & 83.80                         \\
\textbf{SGTK(Our)}    & \cellcolor[HTML]{C4D79B}83.50 & 72.00                         & 80.00                         & 92.87                         & \cellcolor[HTML]{C4D79B}87.90 \\
\textbf{SGNK(Our)}    & 82.60                         & \cellcolor[HTML]{C4D79B}72.40 & \cellcolor[HTML]{C4D79B}80.00 & 92.12                         & 86.50                         \\ \hline
\end{tabular}
\end{table}

\subsection{\textbf{Node Classification Efficiency}}
\begin{figure*}[h]
  \centering
  \subfigure{
    \includegraphics[width=0.18\textwidth]{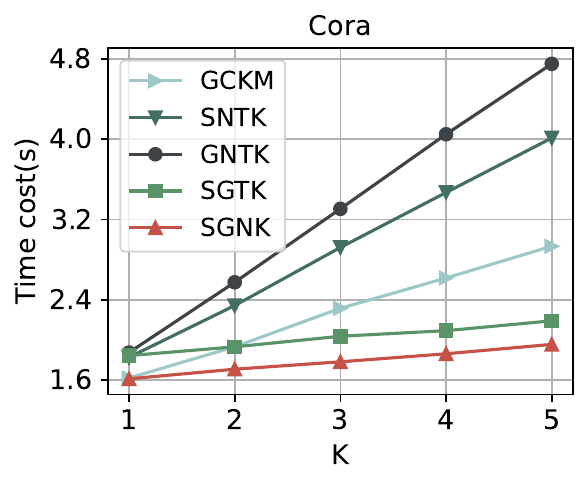}
  }
  \subfigure{
    \includegraphics[width=0.18\textwidth]{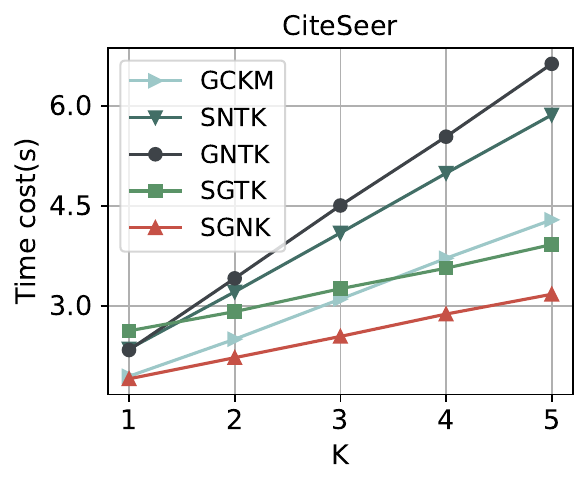}
  }
  \subfigure{
    \includegraphics[width=0.18\textwidth]{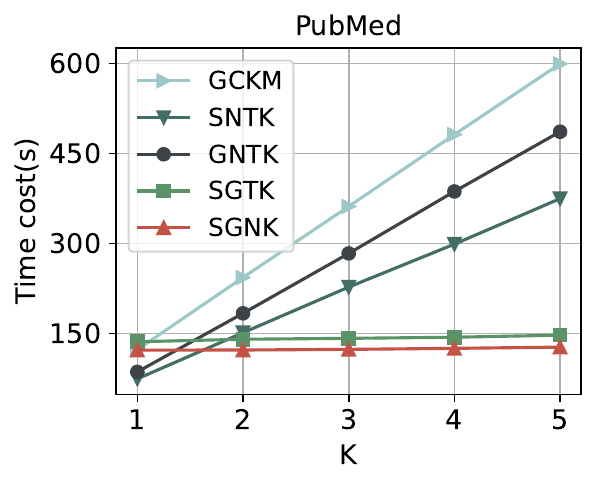}
  }
  \subfigure{
    \includegraphics[width=0.18\textwidth]{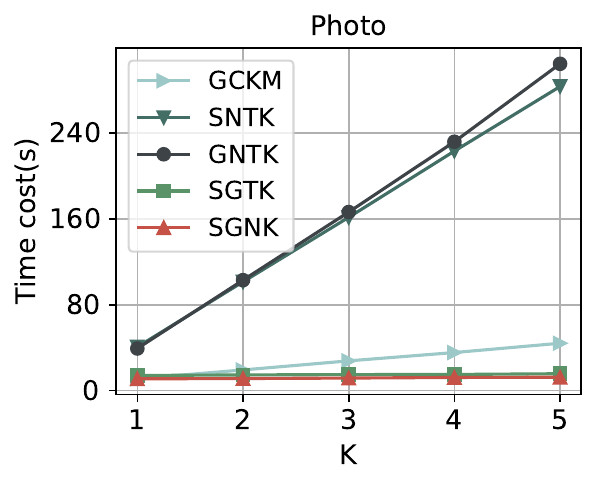}
  }
    \subfigure{
    \includegraphics[width=0.18\textwidth]{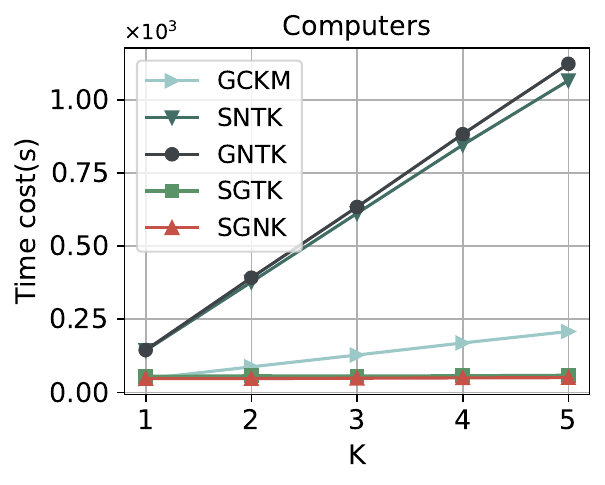}
  }\vspace{-0.5cm} 
  \caption{The changes of node classification time consuming with respect to $K$.}
  \label{fig:node_K_time}
\end{figure*}

To validate the efficiency of the proposed methods in node classification tasks, we investigated the impact of the parameter $K$ on the time consumption of GCKM, SNTK, GNTK, SGTK, and SGNK when calculating the kernel matrix. The experimental results, as shown in Figure~\ref{fig:node_K_time}, indicate that GCKM, GNTK, and SNTK exhibit a linear increase in time consumption with an increase in $K$. In contrast, our methods remain largely unaffected by variations in the message-passing times, maintaining consistent efficiency in kernel matrix computations.

\begin{figure}[h]
  \centering
  \subfigure{
    \includegraphics[width=0.18\textwidth]{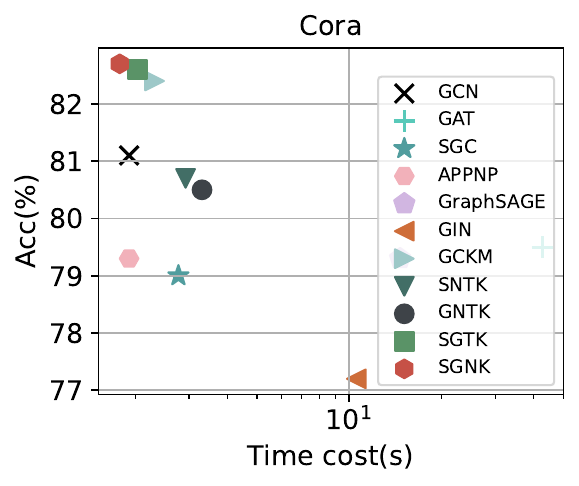}
  }
  \subfigure{
    \includegraphics[width=0.18\textwidth]{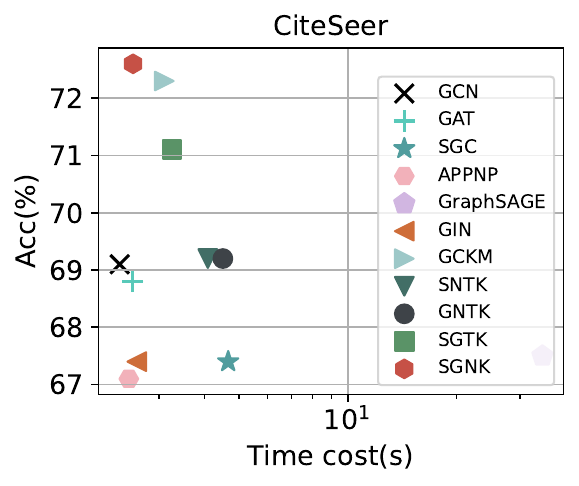}
  }\vspace{-0.4cm} 
  
  \subfigure{
    \includegraphics[width=0.18\textwidth]{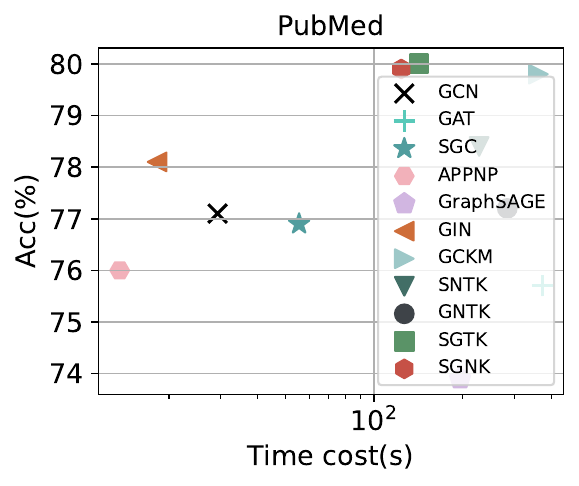}
  }
  \subfigure{
    \includegraphics[width=0.18\textwidth]{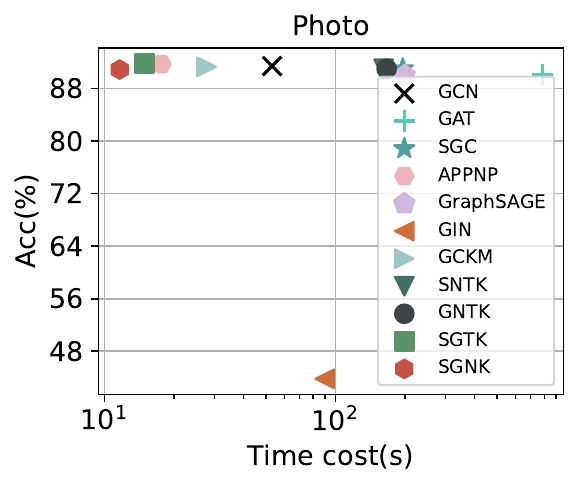}
  }\vspace{-0.4cm} 
  
  \subfigure{
    \includegraphics[width=0.18\textwidth]{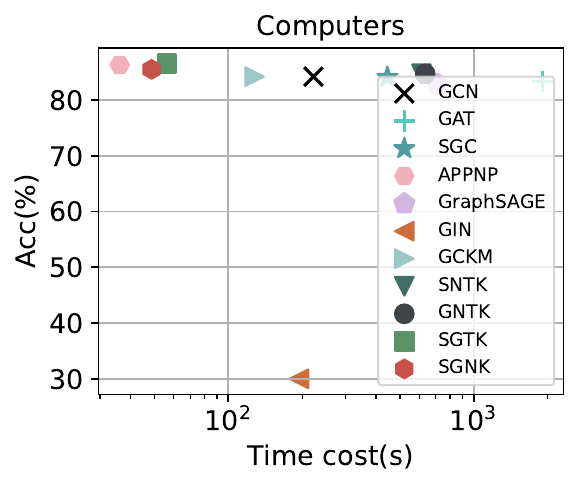}
  }\vspace{-0.5cm} 
  \caption{node classification time consumption and accuracy comparison.}
  \label{fig:node_time}
\end{figure}

Additionally, we present the time taken by each method to achieve the classification accuracy detailed in Table~\ref{tab:node_acc}. This assessment allows us to evaluate the time efficiency of each method. The results of time consumption are presented in Figure~\ref{fig:node_time}. SGNK demonstrates low time consumption across all five datasets, followed closely by SGTK. The proposed methods exhibit time efficiency at least three times faster than GNTK, with SGNK (0.015s) outperforming GNTK (0.827s) by a factor of 53.7 on the Computers dataset. In contrast, GNN methods incur significantly higher time costs due to the need for iterative training over multiple epochs and the uncertainty of model convergence.

In conclusion, the proposed methods not only outperform the baselines in node classification performance but also exhibit consistent time efficiency. This consistency corroborates the theoretical analysis, providing strong evidence for the enhancement in time efficiency achieved by the proposed methods.

\subsection{\textbf{Graph Classification Performance Comparison}}
This subsection evaluates the proposed kernel methods for modeling similarity between graphs. To this end, we conduct experiments on four graph-level datasets, which include social network and molecular graph datasets. We employ the SVM classifier with 10-fold cross-validation to assess the performance of the proposed graph kernel methods. The evaluation metrics include the average classification accuracy and standard deviation, which are reported to provide a quantitative measure of the methods' efficacy.

As shown in Table~\ref{tab:graph_acc}, the proposed SGTK method achieves competitive classification performance on three out of the four datasets: IMDB-BINARY, IMDB-MULTI, and PTC\_MR. Compared to other graph kernel methods, SGTK demonstrates consistent improvements in accuracy on these datasets. It is worth noting that the proposed method does not involve a learning or training phase, yet it achieves results comparable to—and, in some cases, better than—GNN methods that require extensive training.

The classification accuracy on graph datasets demonstrates the proposed method's effectiveness in measuring graph similarity. Despite their lower computational complexity, the proposed kernel methods deliver competitive performance, making them a practical choice for applications where computational efficiency is critical.

\begin{table}[]
\caption{Graph classification accuracy (\%) and standard derivation. The top three performance methods are highlighted in varying shades of green, with darker shades representing superior performance.}
\label{tab:graph_acc}
\begin{tabular}{ccccc}
\hline
Methods             & IMDB-B                           & IMDB-M                                                  & PTC                              & MUTAG                            \\ \hline
GCN                 & 74.0±3.4                         & 51.9±3.8                                                & 64.2±4.3                         & 85.6±5.8                         \\
GAT                 & 75.8±3.4                         & 52.0±3.5                                                & 64.3±7.5                         & 81.1±7.2                         \\
SGC                 & 69.6±4.9                         & 47.4±2.0                                                & 58.7±6.5                         & 66.5±9.6                         \\
APPNP               & 75.2±3.4 & 52.1±3.1                                                & 62.0±7.9                         & 78.3±6.5                         \\
GraphSAGE           & 72.3±5.3                         & 50.9±2.2                                                & 63.9±7.7                         & 85.1±7.6                         \\
GIN                 & 75.1±5.1                         & 52.3±2.8                                                & 64.6±7.0 & 89.4±5.6 \\
WL subtree          & 73.8±3.9                         & 50.9±3.8                                                & 59.9±4.3                         & \cellcolor[HTML]{C4D79B}90.4±5.7 \\
RetGK               & 71.9±1.0                         & 47.7±0.3                                                & 62.5±1.6                         & \cellcolor[HTML]{D8E4BC}90.3±1.1 \\
GraphQNTK           & 73.3±3.6                         & 48.1±4.3                                                & 62.9±5.0                         & 88.4±6.5                         \\
GCKM                & 75.4±2.4                         & \cellcolor[HTML]{C4D79B}53.9±2.8                        & \cellcolor[HTML]{C4D79B}67.7±5.4 & 88.7±7.6                         \\
SNTK                & {\color[HTML]{333333} 75.3±2.4}  & \cellcolor[HTML]{EBF1DE}{\color[HTML]{333333} 52.9±4.5} & {\color[HTML]{333333} 65.6±5.1}  & 85.6±6.3                         \\
GNTK                & \cellcolor[HTML]{D8E4BC}76.5±2.8                         & 52.8±4.2                                                & \cellcolor[HTML]{D8E4BC}67.7±5.7                         & \cellcolor[HTML]{EBF1DE}90.0±8.4                         \\
\textbf{SGTK(Ours)} & \cellcolor[HTML]{C4D79B}76.6±3.5 & \cellcolor[HTML]{D8E4BC}52.9±3.8                        & \cellcolor[HTML]{EBF1DE}66.5±8.5 & 85.7±6.6                         \\
\textbf{SGNK(Ours)} & \cellcolor[HTML]{EBF1DE}75.5±3.8 & 52.3±4.5                                                & 63.5±7.6                         & 86.7±6.2                         \\ \hline
\end{tabular}
\end{table}


\subsection{\textbf{Graph Classification Efficiency}}

The subsection empirically evaluates the computational efficiency of the proposed methods on graph datasets, as outlined in the theoretical complexity analysis. In the experiments, we vary the parameter $K$ within the range ${1, 2, 3, 4, 5}$ and measure the time required to compute kernel matrices across five graph-level datasets.

\begin{figure}[]
  \centering
  \subfigure{
    \includegraphics[width=0.18\textwidth]{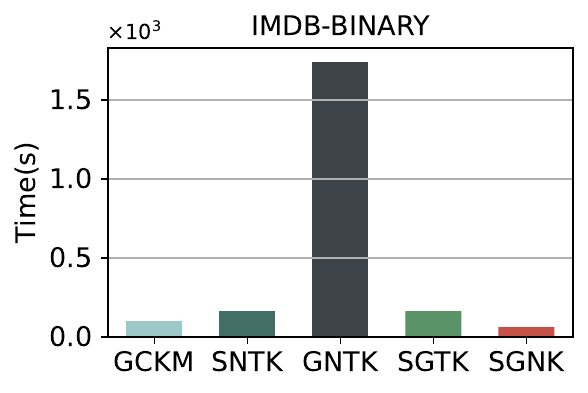}
  }
  \subfigure{
    \includegraphics[width=0.175\textwidth]{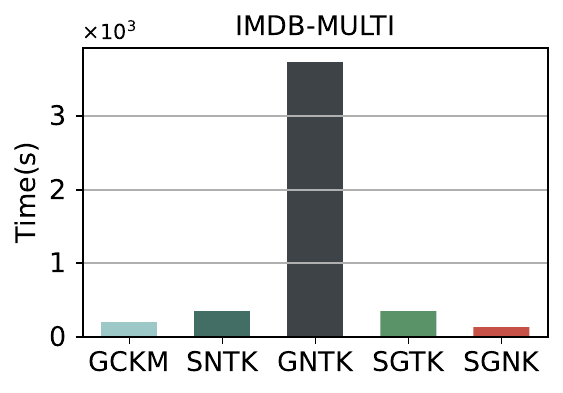}
  }
  
  \subfigure{
    \includegraphics[width=0.18\textwidth]{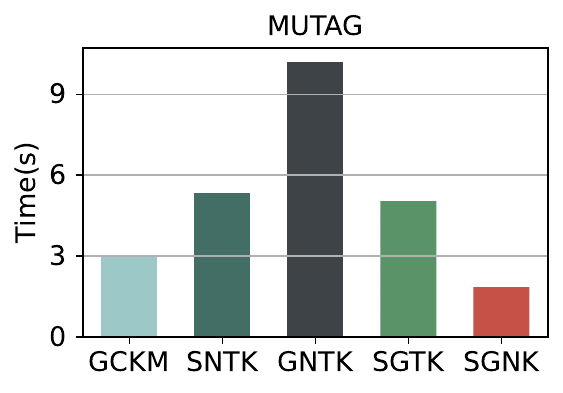}
  }
  \subfigure{
    \includegraphics[width=0.18\textwidth]{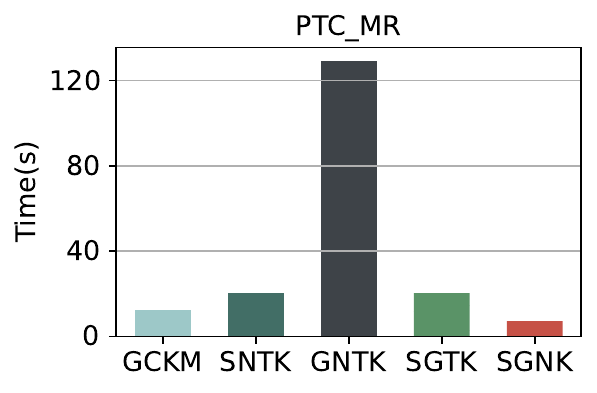}
  }\vspace{-0.5cm} 
  \caption{Graph classification time consuming comparison.}
  \label{fig:graph_time}
\end{figure}

Figure~\ref{fig:graph_time} presents the computation time for the GCKM, SNTK, GNTK, SGTK, and SGNK methods, corresponding to the classification accuracies reported in Table~\ref{tab:graph_acc}. The results align with the theoretical complexity analysis: SGNK is the most efficient method, while GNTK requires the longest computation time. Notably, the proposed methods, SGTK and SGNK, achieve at least a fourfold reduction in computation time compared to GNTK.

\begin{figure}[h]
  \centering
  \subfigure{
    \includegraphics[width=0.178\textwidth]{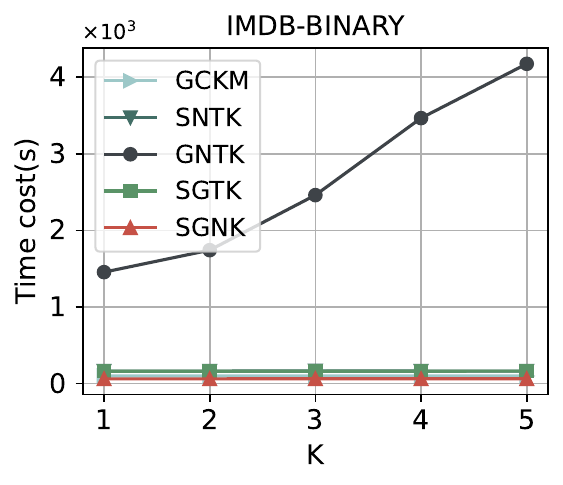}
  }
  \subfigure{
    \includegraphics[width=0.185\textwidth]{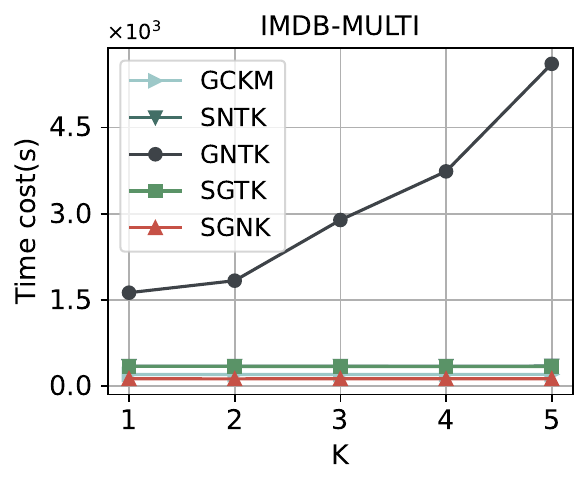}
  }
  
  \subfigure{
    \includegraphics[width=0.184\textwidth]{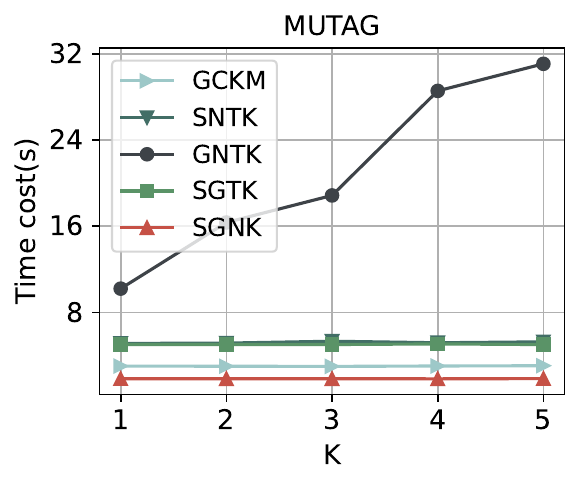}
  }
  \subfigure{
    \includegraphics[width=0.19\textwidth]{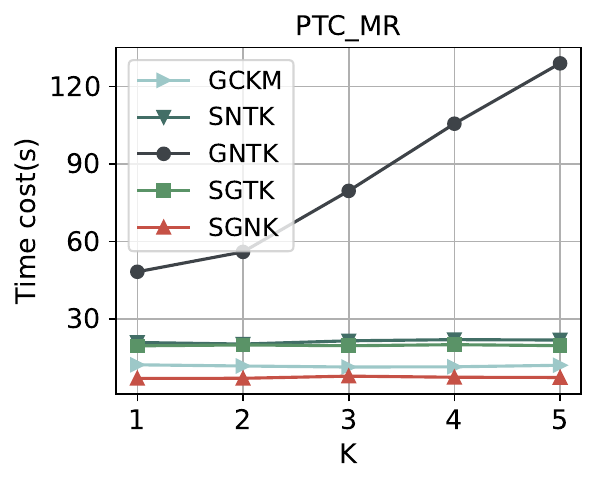}
  }\vspace{-0.5cm} 
  \caption{Variations in graph classification time consumption with changes in $K$.}
  \label{fig:graph_K_time}
\end{figure}

To investigate the impact of the $K$ parameter on computation time, we analyze the relationship between kernel matrix computation time and $K$. Table~\ref{tab:graph_K_time} summarizes the time consumption data for all methods, and Figure~\ref{fig:graph_K_time} illustrates the results for comparison.
As shown in Figure~\ref{fig:graph_K_time}, the computation time of GNTK increases linearly with $K$. In contrast, the proposed methods, SGTK and SGNK, which employ continuous $K$-fold aggregation, exhibit stable computation times and achieve the lowest time costs across all tested $K$ values.

In summary, the experimental results demonstrate the computational efficiency of the proposed SGTK and SGNK methods, providing strong empirical support for the theoretical complexity analysis. These methods offer a practical advantage in scenarios where computational efficiency is critical.

\begin{table}[h]
\caption{Time required (s) to calculate the graph-level kernel matrix as varying $K$.}
\label{tab:graph_K_time}
\centering
\setlength{\tabcolsep}{3pt} 
\begin{tabular}{ccccccc}
\hline
Dataset                     & K & GCKM     & SNTK     & \multicolumn{1}{c}{GNTK} & SGTK     & SGNK     \\ \hline
\multirow{5}{*}{IMDB-B} & 1 & 99.1796  & 162.5266 & 1455.04                  & 161.0031 & 61.5029  \\
                            & 2 & 98.7254  & 162.4243 & 1742.01                  & 160.7982 & 61.5988  \\
                            & 3 & 99.4667  & 164.5200 & 2460.97                  & 159.5090 & 61.8451  \\
                            & 4 & 98.1082  & 162.8410 & 3467.13                  & 160.4228 & 61.8403  \\
                            & 5 & 98.6913  & 161.8836 & 4176.06                  & 162.3239 & 61.8523  \\ \hline
\multirow{5}{*}{IMDB-M}  & 1 & 203.5473 & 346.7882 & 1626.47                  & 343.5592 & 128.5789 \\
                            & 2 & 202.5733 & 346.9627 & 1835.87                  & 343.5564 & 127.9949 \\
                            & 3 & 203.2251 & 349.2096 & 2891.54                  & 342.3580 & 129.7378 \\
                            & 4 & 204.3999 & 346.9047 & 3737.56                  & 340.1037 & 128.3960 \\
                            & 5 & 204.7037 & 350.6760 & 5609.25                  & 341.6731 & 128.6607 \\ \hline
\multirow{5}{*}{MUTAG}      & 1 & 3.0228   & 5.1242   & 10.21                    & 5.0305   & 1.8515   \\
                            & 2 & 3.0084   & 5.1546   & 16.35                    & 5.0241   & 1.8481   \\
                            & 3 & 2.9935   & 5.3291   & 18.86                    & 5.0208   & 1.8487   \\
                            & 4 & 3.0312   & 5.1870   & 28.56                    & 5.0793   & 1.8402   \\
                            & 5 & 3.0546   & 5.2734   & 31.07                    & 5.0176   & 1.8589   \\ \hline
\multirow{5}{*}{PTC}        & 1 & 12.2249  & 20.8693  & 48.25                    & 19.5684  & 6.9742   \\
                            & 2 & 11.7598  & 20.2613  & 55.96                    & 19.9208  & 7.0001   \\
                            & 3 & 11.3650  & 21.5222  & 79.59                    & 19.6073  & 7.8397   \\
                            & 4 & 11.4551  & 21.9708  & 105.60                   & 20.0212  & 7.4049   \\
                            & 5 & 12.0432  & 21.8219  & 128.97                   & 19.6431  & 7.3243   \\ \hline
\end{tabular}
\end{table}

\begin{figure}[h]
  \centering
  \subfigure{
    \includegraphics[width=0.18\textwidth]{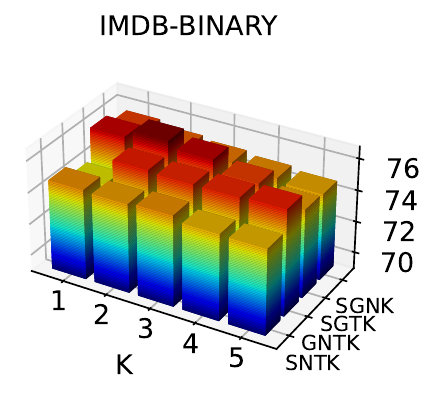}
  }
  \subfigure{
    \includegraphics[width=0.18\textwidth]{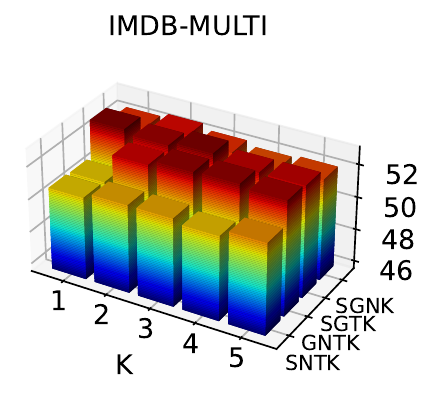}
  }\vspace{-0.5cm} 
  
  \subfigure{
    \includegraphics[width=0.18\textwidth]{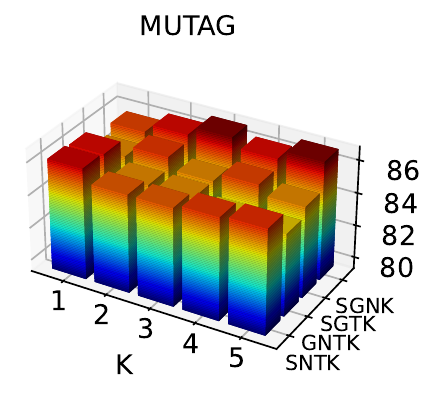}
  }
  \subfigure{
    \includegraphics[width=0.18\textwidth]{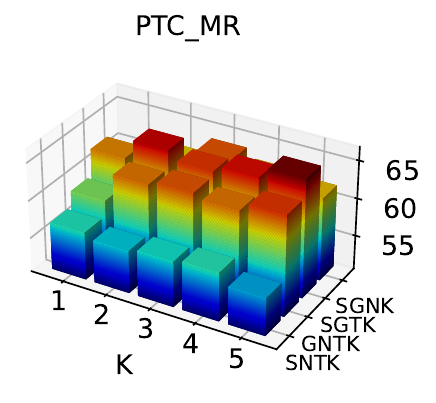}
  }\vspace{-0.5cm} 
  \caption{The changes of graph classification accuracy with respect to $K$. The height of the column, representing classification accuracy, directly correlates with the color at the top, with higher columns indicated by darker colors.}\label{fig:graph_K_acc}
\end{figure}

\begin{figure}[htb]
  \centering
  \subfigure{
    \includegraphics[width=0.18\textwidth]{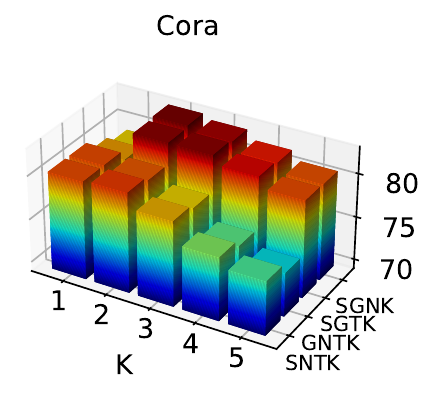}
  }
    \subfigure{
    \includegraphics[width=0.18\textwidth]{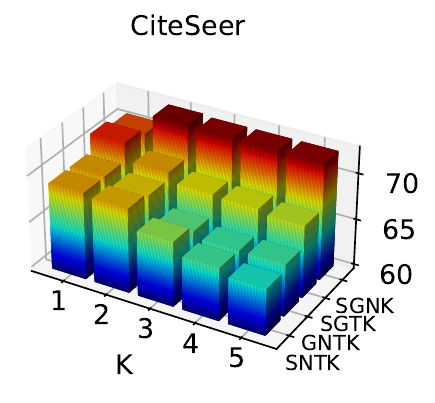}
  }\vspace{-0.5cm} 
  
  \subfigure{
    \includegraphics[width=0.18\textwidth]{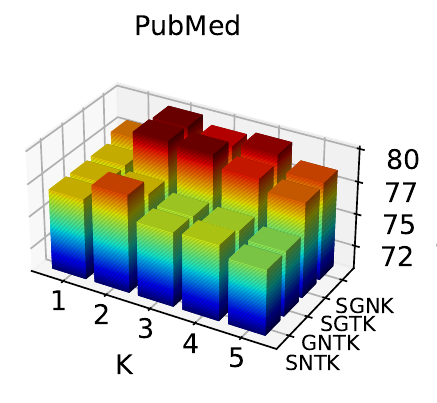}
  }
  \subfigure{
    \includegraphics[width=0.18\textwidth]{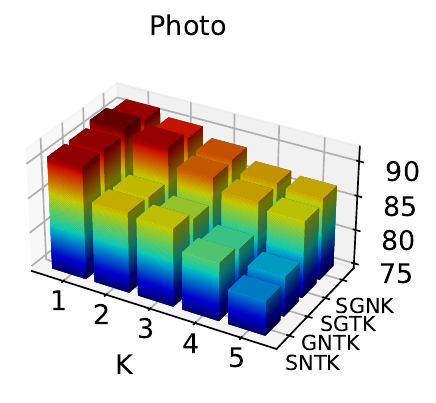}
  }\vspace{-0.5cm} 
  
  \subfigure{
    \includegraphics[width=0.18\textwidth]{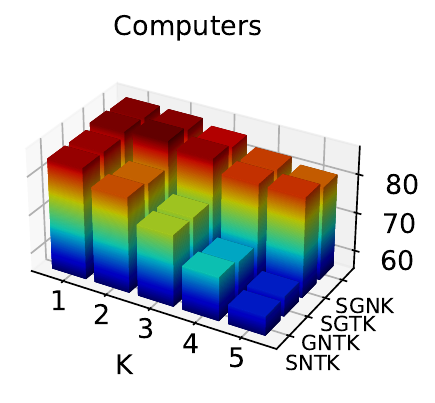}
  }\vspace{-0.5cm} 
    \caption{The changes of node classification accuracy with respect to $K$. The height of the column, representing classification accuracy, directly correlates with the color at the top, with higher columns indicated by darker colors.}\label{fig:K_acc}
\end{figure}

\subsection{\textbf{Parameters Sensitive Analysis}}
This subsection investigates the impact of different  $K$ parameter values on classification accuracy in both graph and node classification tasks. We varied $K$ within the range \{1, 2, 3, 4, 5\} and reported the changes in classification accuracy for SNTK, GNTK, SGTK, and SGNK on four graph classification datasets and five node classification datasets, as depicted in Figure~\ref{fig:graph_K_acc} and Figure~\ref{fig:K_acc}.

For the graph classification task, the four methods demonstrate varying degrees of fluctuation in classification accuracy on the IMDB-MULTI and PTC\_MR datasets. In contrast, the proposed methods, SGTK and SGNK, exhibit more stable classification performance across different $K$ values on the four datasets. This observation underscores the dataset-dependent behavior of these methods. 
In the node classification task, SGTK and SGNK demonstrate minimal sensitivity to changes in $K$, maintaining stable classification accuracy across different datasets. In contrast, GNTK shows notable variability in classification accuracy when $K$ changes, particularly on the Cora, CiteSeer, Photo, and Computers datasets. This contrast highlights the robustness of SGTK and SGNK compared to GNTK under varying $K$ values.

These results reveal the methods' differing sensitivities to the aggregation parameter $K$ across different datasets. This underscores the importance of parameter selection for achieving stable and high classification accuracy in both graph and node classification tasks.

\section{Conclusion}
\label{sec:conclusion}

In this paper, we propose the Simplified Graph Neural Tangent Kernel (SGTK) and its derivative, the Simplified Graph Neural Kernel (SGNK), to address the computational inefficiencies of graph kernels. By introducing continuous $K$-step aggregation and eliminating multi-block stacking, SGTK reduces redundancy in the Graph Neural Tangent Kernel (GNTK). SGNK further simplifies computations by interpreting an infinite-width simple graph neural network as a Gaussian process.
Experimental results demonstrate that SGTK and SGNK achieve competitive classification accuracy while improving computational efficiency. These methods provide practical and scalable alternatives to traditional graph kernels and graph neural networks, particularly in resource-constrained scenarios.


\balance
\bibliographystyle{ACM-Reference-Format}
\bibliography{references}

\end{document}